\documentclass[letterpaper, 10 pt, conference]{ieeeconf}  

\IEEEoverridecommandlockouts                              

\usepackage{graphics} 
\usepackage{epsfig} 
\usepackage{times} 
\usepackage{amsmath} 
\usepackage{amssymb}  
\usepackage{mathrsfs}
\usepackage{multirow} 
\usepackage{subfigure}
\def\ie{{\em i.e.}}
\def\eg{{\em e.g.}}
\def\etal{{\em et. al.}}

\title{Learning Local Feature Descriptor with Motion Attribute \\ For Vision-based Localization}

\author{
Yafei Song\textsuperscript{12*}, Di Zhu\textsuperscript{1}, Jia Li\textsuperscript{3}, Yonghong Tian\textsuperscript{2} and Mingyang Li\textsuperscript{1} 
\thanks{\textsuperscript{1}Yafei Song, Di Zhu and Mingyang Li are with the AI Labs, Alibaba Group, Hangzhou 311121, China. {\tt\small \{huaizhang.syf, david.zd, mingyangli\}@alibaba-inc.com}}%
\thanks{\textsuperscript{2}Yafei Song and Yonghong Tian are with the National Engineering Laboratory for Video Technology, School of Electronics Engineering and Computer Science, Peking University, Beijing 100871, China. {\tt\small \{songyf, yhtian\}@pku.edu.cn}}%
\thanks{\textsuperscript{3}Jia Li is with the the National Key Laboratory of Virtual Reality Technology and System, School of Computer Science and Engineering, Beihang University, Beijing 100191, China. {\tt\small \{jiali\}@buaa.edu.cn}}%
\thanks{\textsuperscript{*}Corresponding author.}%
}

\begin{document}

\maketitle
\thispagestyle{empty}
\pagestyle{empty}

\begin{abstract}
In recent years, camera-based localization has been widely used for robotic applications, and most proposed algorithms rely on local features extracted from recorded images. For better performance, the features used for open-loop localization are required to be short-term globally static, and the ones used for re-localization or loop closure detection need to be long-term static. Therefore, the motion attribute of a local feature point could be exploited to improve localization performance, \eg, the feature points extracted from moving persons or vehicles can be excluded from these systems due to their unsteadiness. In this paper, we design a fully convolutional network (FCN), named MD-Net, to perform motion attribute estimation and feature description simultaneously. MD-Net has a shared backbone network to extract features from the input image and two network branches to complete each sub-task. 
With MD-Net, we can obtain the motion attribute while avoiding increasing much more computation. 
Experimental results demonstrate that the proposed method can learn distinct local feature descriptor along with motion attribute only using an FCN, by outperforming competing methods by a wide margin. We also show that the proposed algorithm can be integrated into a vision-based localization algorithm to improve estimation accuracy significantly.
\end{abstract}

\section{Introduction}

Image local features are important for many tasks and applications, \eg, image-based localization \cite{Song2016TMM, li2014high, li2013high, concha2015dpptam}, structure from motion \cite{Huang_2018_ICRA, Zhu_2018_CVPR} and simultaneous localization and mapping \cite{Mur_2017_orb_slam2, schneider2018maplab,li2013optimization}. 
Previous researchers have designed various and powerful local feature detectors and descriptors, \eg,  SIFT \cite{SIFT_2004_ijcv}, ORB \cite{ORB_2011_iccv} and FREAK \cite{FREAK_2012_cvpr}. 
Due to the recent success of deep neural networks on various tasks \cite{Ren_2017_faster_rcnn, Shelhamer_2017_FCN}, several previous papers have attempted to learn robust image local features automatically \cite{TILDE_2015_cvpr, LIFT_2016_eccv, Tian_2017_cvpr, Mishchuk_2017_nips, superpoint_2018_cvpr}.

\begin{figure}
  \center
  \includegraphics[width=1\columnwidth]{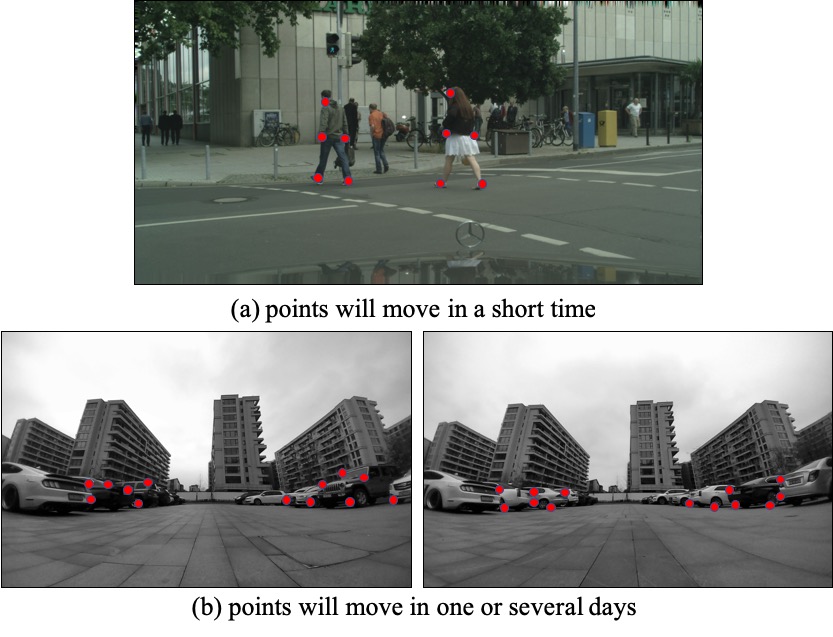}
  \caption{The moving points are ubiquitous in the images, including short term moving points such as in (a) and long term moving points such as in (b). All these points will decrease the performance of a localization system.}
  \label{fig:teaser}
\end{figure}

For a vision-based localization system, the motion attribute of each feature point could be exploited to improve the performance, \eg, as demonstrated in Fig. \ref{fig:teaser}(a), the points on moving people should be excluded. 
Existing systems usually perform this step by eliminating the point matches via epipolar geometry constraints \cite{MVG_2004_book}. 
These methods can successfully remove points corresponding to fast-moving objects but have difficulties to block points corresponding to slow-moving objects.
In addition, this strategy will fail on long-term moving points, \eg, the points on the parked vehicles, as shown in Fig. \ref{fig:teaser}(b). Such points will be static in a short time but will move to another place in a few minutes or even days. 
This will inevitably lead to reduced performance for a couple of localization operations (e.g., loop closure detection or re-localization\cite{schneider2018maplab,lynen2015get,zhang2019large}).
Beyond that, there are also some points in unstable areas, such as the sky. In this paper, we focus on selecting the long term static points for localization systems by estimating the motion attribute of each point.

Some previous methods have also exploited the motion attribute to perform localization or mapping. 
Kaneko \etal~\cite{mask_slam_2018_cvpr} first segmented the input image and then only used the points with long term static motion attribute derived from the semantic label. 
However, semantic segmentation is a difficult task, which will dramatically increase the computational cost. 
In this paper, we transform the task to estimate the motion attribute for each feature point, which is easier than semantic segmentation but is sufficient for vision based localization. 
Naseer \etal~\cite{Naseer_2017_icra} exploited geometrically robust regions to determine the position of the input image, which is a different task from localization.

Motion attribute estimation is generally considered as a highly computational costly task for a robotic platform, and thus efficiency is of significant importance for algorithm design. 
Inspired by the recent progress on local feature learning \cite{LIFT_2016_eccv, Tian_2017_cvpr, Mishchuk_2017_nips, superpoint_2018_cvpr}, we try to use a single network to perform the descriptor calculation and the motion attribute estimation tasks simultaneously, since these tasks can be formulated into similar computation processes. 
A lot of existing methods, \eg, \cite{LIFT_2016_eccv, Tian_2017_cvpr, Mishchuk_2017_nips}, focused on extracting distinct features from an image patch. 
However, it is difficult to estimate the motion attribute only from a local patch since the context is essential. Moreover, patch-based algorithms usually have poorer computational efficiency than whole image-based algorithms due to the repetitive computation between different patches, \eg, \cite{rcnn_2014_cvpr} and \cite{Ren_2017_faster_rcnn}.

With the above analysis in mind, we design a fully convolutional network, named MD-Net, which takes the whole image as input and performs motion attribute estimation and feature description simultaneously. 
MD-Net has a shared backbone network, named $\mathcal{N}_{\text{B}}$, and two network branches, named $\mathcal{N}_{\text{M}}$ and $\mathcal{N}_{\text{D}}$. 
The heavy backbone network $\mathcal{N}_{\text{B}}$ takes the whole image as input and outputs shared features to each following network branch. 
The motion attribute estimation branch $\mathcal{N}_{\text{M}}$ then assigns each point as unstable, moving or static. 
Finally, the feature description branch $\mathcal{N}_{\text{D}}$ extracts distinctive descriptor for each point. 

The most similar methods with this paper are from DeTone \etal, \ie, \cite{superpoint_2018_cvpr} and \cite{superpoint_v2_2018_arxiv}. 
In \cite{superpoint_2018_cvpr}, DeTone \etal~have designed an FCN \cite{Shelhamer_2017_FCN} based single network to perform local feature localization and description simultaneously. 
However, the localized points are not robust enough due to the poor localization ability of the standard convolutional neural network (CNNs) \cite{Szegedy_2013_nips}. 
Therefore, we resort to hand-crafted detectors in this paper. 
Moreover, the strategies to learn descriptor are also different. 
In \cite{superpoint_v2_2018_arxiv}, the model also estimates whether a point is steady. However, this work can only handle short term moving points as shown in Fig. \ref{fig:teaser}(a), but not long term moving points as shown in Fig. \ref{fig:teaser}(b). 
We also experimentally show that, when integrated into vision-based localization, the proposed algorithm is able to achieve significantly better accuracy, compared to \cite{superpoint_v2_2018_arxiv}.
 
The contributions of this paper mainly lie in four aspects:
\begin{enumerate}
\item We design an FCN to estimate the motion attribute of each local feature point to distinguish static points from moving or unstable points. 
\item We further enhance the FCN to calculate the descriptor of each point simultaneously, which is more efficient than patch-based methods.
\item We integrate the proposed local feature processing pipeline into a vision-based localization method to seek accuracy gain.
\item Experimental results demonstrate that both the proposed local feature method and integrated localization algorithm outperform competing state-of-the-art algorithms by a wide margin.
\end{enumerate}

\section{Related Work}

A local feature algorithm usually can be divided into two steps: feature localization and feature descriptor calculation. In this section, we briefly introduce some well-known local features, including hand-crafted and learning-based, from these two aspects.

\textbf{Hand-crafted local features}. 
Over the last few decades, researchers have designed various algorithms to localize each robust point as well as calculate its distinct descriptor. 
At the very beginning, many methods devoted to detecting and localizing corners in the image, \eg, Harris corner \cite{Harris_1988_AVC}, FAST \cite{FAST_2006_eccv}. 
Besides corner detection methods, researchers also introduced the scale-space theory and detected the extrema as feature points, \eg, Laplacian of Gaussian (LoG) \cite{Lindeberg_1998_ijcv}, difference of Gaussians (DoG) \cite{SIFT_2004_ijcv} and MSER \cite{MSER_2002_bmvc}. 
Researchers have also designed various local feature descriptors, \eg, SIFT \cite{SIFT_2004_ijcv}, SURF \cite{SURF_2006_eccv}, and HOG \cite{Zhu_2006_cvpr}. 
As robotics tasks usually need real-time algorithms, some fast binary descriptors also have been proposed, \eg, BRIEF \cite{brief_2010_eccv}, FREAK \cite{FREAK_2012_cvpr}, ORB \cite{ORB_2011_iccv}. 
These local feature detectors and descriptors have been successfully applied in many tasks. 
However, with the successful applications of deep learning based methods on various tasks, researchers have attempted to learn more robust local features from the data automatically. 

\textbf{Learning-based local features}. 
Learning based methods can be divided into three classes. 
The first class methods only learn robust detectors \cite{TILDE_2015_cvpr, Savinov2017, Zhang2017}. 
One challenge is how to generate ground truth. To this end, Verdie \etal~\cite{TILDE_2015_cvpr} applied DoG algorithm on all images of a unique scene and selected the points which can be detected in most cases. 
This strategy can enable the learned detector to outperform the baseline detector. 
Savinov \etal~\cite{Savinov2017} formulated and solved the problem in an unsupervised manner. 
Zhang \etal \cite{Zhang2017} focused on learning a discriminative and transformation covariant detector. 
These detectors cannot remove the points on moving objects.

The second class methods aim at learning robust descriptors. 
To this end, several large-scale datasets \cite{Brown_2007_ijcv, HPatches_2017_cvpr, PS_Dataset_2018_arxiv} are collected for training and evaluation. 
Tian \etal~\cite{Tian_2017_cvpr} designed a convolutional neural network named L2-Net to extract descriptors from a local patch. 
Mishchuk \etal~\cite{Mishchuk_2017_nips} further improved the performance with a large margin via minimizing the hard negative samples. 
Luo \etal~\cite{Luo2018} exploited geometric constraints to learn more robust descriptors. 
However, these methods calculate descriptor for each image patch, which may be not efficient than FCN based algorithms.

Yi \etal~\cite{LIFT_2016_eccv} and Ono \etal~\cite{Ono2018} divided the whole process into several steps as previous algorithms and performed each step using a neural network. 
These methods can take full advantage of previous knowledge on the problem but not efficient than FCN based methods. 
DeTone \etal~\cite{superpoint_2018_cvpr} designed an FCN based single network to perform local feature localization and description simultaneously. 
However, these methods are not designed for localization and cannot remove moving points. 
To this end, DeTone \etal~\cite{superpoint_v2_2018_arxiv} further estimated whether a point is steady and only selected the static points. 
However, this work can only handle short term moving points but not long term moving points.

\begin{figure*}
\center
  \includegraphics[width=16cm]{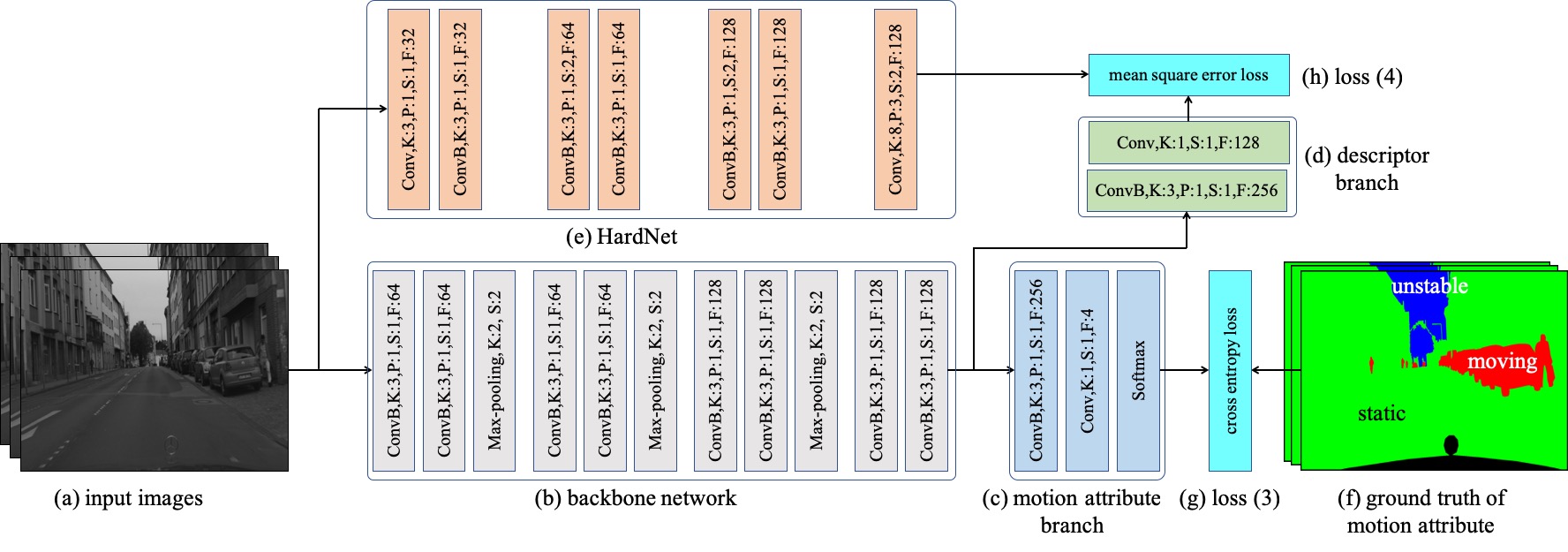}
  \caption{The training process for our MD-Net. We feed a mini-batch of training samples into the backbone network. Subsequently, the extracted features are feed into two branches simultaneously. For motion attribute, we use cross-entropy loss \eqref{eq:loss:motion:weighted}, and for descriptor, we use mean square error loss \eqref{eq:loss:desc}. Note that, Conv denotes a convolutional layer, ConvB denotes that it is followed by a batch normalization layer and a ReLU layer, K denotes kernel size, P denotes padding size, S denotes stride size, and F denotes the number of output feature maps.}
  \label{fig:pipeline}
\end{figure*}

\section{Learning Descriptor with Motion Attribute}

In this section, we first briefly overview the whole method and detailedly introduce each module subsequently. As demonstrated in Fig. \ref{fig:pipeline}, we design an FCN named MD-Net to estimate motion attribute and calculate descriptor for each point in the input grayscale image. The MD-Net consists of a heavy backbone network named $\mathcal{N}_{\text{B}}$ to extract shared features from the input image, a light branch $\mathcal{N}_{\text{M}}$ to estimate motion attribute, and another light branch $\mathcal{N}_{\text{D}}$ to calculate descriptors. For a fair comparison, the backbone network $\mathcal{N}_{\text{B}}$ has the same structure with the backbone network in \cite{superpoint_2018_cvpr}, which consists of $8$ convolutional layers and $3$ max-pooling layers. Each convolutional layer is followed up with a batch normalization layer \cite{Ioffe_Szegedy_2015_batchnorm} and a ReLU activation layer \cite{Nair_Hinton_2010_relu}. The hyper-parameters of each layer also can be found in Fig. \ref{fig:pipeline}(b). The backbone network $\mathcal{N}_{\text{B}}$ outputs a group of feature maps to the following branches. This structure has high computational efficiency benefiting from the fully convolutional structure and the down-sampling pooling layer.

\subsection{Motion Attribute Estimation}

A local feature detector usually localizes a large number of points in an image. However, not all of these points are suitable for localization or mapping. In this paper, we divide the motion attribute into three classes, \ie, $\mathcal{A} = \{ \mathcal{A}_{\text{u}}, \mathcal{A}_{\text{m}}, \mathcal{A}_{\text{s}} \}$, where $\mathcal{A}_{\text{u}}$ denotes unstable, $\mathcal{A}_{\text{m}}$ denotes moving, $\mathcal{A}_{\text{s}}$ denotes static. The unstable points are on the image regions, which will vary fast, such as the cloud in the sky. The moving points are on the moving objects, including short term and long term. The static points are steady for a long time. A localization system usually can automatically discard short-term fast-moving points as these points could not meet the epipolar constraint.
However, when it comes to very-slowly-moving points, it becomes more challenging. 
Moreover, the long-term moving points can not be removed, which will result in decreased performance, even with a self-improving method \cite{superpoint_v2_2018_arxiv}. When localizing an input image, all of the unstable and moving points will decrease the performance. To alleviate this phenomenon, we propose to assign a point's motion attribute according to its semantic label. As previous works have constructed many semantic segmentation datasets, \eg, Cityscapes \cite{Cordts2016Cityscapes}, we only need to perform a label transforming process.

To estimate the motion attribute, we add a network branch $\mathcal{N}_{\text{M}}$ following the backbone network, as shown in Fig. \ref{fig:pipeline}, which consists of two convolutional layers, one batch normalization layer, one ReLU layer and one softmax layer.  This branch can transform the input features of each point to the probability of each motion attribute. The hyperparameters of this branch can be found in Fig. \ref{fig:pipeline}(c). To train this branch, we adopt the cross-entropy loss 
\begin{equation}
\label{eq:loss:motion}
\mathcal{L}_{\text{M}} = - \sum_{i,j} y_{ij} \log(p_{ij}),
\end{equation}
where $y_{ij} = 1$ only if the motion attribute of point $i$ is $\mathcal{A}_{j}$, otherwise $y_{ij} = 0$, $p_{ij}$ is the predicted probability.

The cross-entropy loss can effectively supervise a network when the distribution of sample sizes of all classes is uniform. Otherwise, the model will incline to the class with large size \cite{Li_2018_arxiv}. This unbalance bias is ubiquitous in the training data. To this end, we re-weight the loss of each class according to its size as 
\begin{equation}
\label{eq:weight:motion}
w_{j} = \frac{\frac{1}{N_{j}}}{\sum_{j} \frac{1}{N_{j}}},
\end{equation}
where $N_{j}$ is the sample size of class ${\mathcal{A}_j}$. Then the loss function \eqref{eq:loss:motion} can be transformed to its re-weighted version as
\begin{equation}
\label{eq:loss:motion:weighted}
\mathcal{L}_{\text{M}} = - \sum_{i,j} w_{j} y_{ij} \log(p_{ij}).
\end{equation}
With the re-weighting strategy, each class would contribute equally to the loss function to avoid the influence of unbalance classes distribution.

\subsection{Descriptor Calculation}

To learn distinct descriptors from the data automatically, previous methods have constructed several elaborate datasets, \eg, \cite{Brown_2007_ijcv, HPatches_2017_cvpr, PS_Dataset_2018_arxiv}. As these datasets only have image patches, most methods also based on patches \cite{Tian_2017_cvpr, Mishchuk_2017_nips, Luo2018}. For deep learning algorithms, however, image-based models usually are more efficient than patch-based models. If the network structure is unique, the computational cost has a  linear relationship with the size of the input image. Previous methods typically take a $32 \times 32$ patch as input. Its computation cost will be equal to a whole image-based model if the input image is $320 \times 320$ and the detector only detects $100$ points. In practical, there are often more than $300$ points in an image. In other words, a patch-based model usually need several times of computational cost compared with the corresponding whole image-based model.

In this paper, we design an FCN network to calculate feature descriptor from the whole input image as in Fig. \ref{fig:pipeline}(a)(b)(d). However, it is a challenge to design an effective training strategy. DeTone \etal~\cite{superpoint_2018_cvpr} directly used all points in the image. However, this strategy will introduce plenty of noisy samples. To take full advantage of existing datasets, we resort to a teacher-student framework, which is usually exploited for feature embedding \cite{Song_2017_iccv}. Without loss of generality, we transform the HardNet model \cite{Mishchuk_2017_nips} to an image-based model and take it as the teacher model, which can be used to supervise the descriptor learning. As shown in Fig. \ref{fig:pipeline}(e), we can see the detailed structure of the modified HardNet. Compared with the initial version, we only remove the last reshape layer and add padding on the last convolutional layer. This network surgery has little influence on the effectiveness of the model.

To train the descriptor branch $\mathcal{N}_{\text{D}}$, which is the student model, we minimize the mean square error between the outputs of $\mathcal{N}_{\text{D}}$ and HardNet. The loss function can be defined as
\begin{equation}
\label{eq:loss:desc}
\mathcal{L}_{\text{D}} = \sum_{i,k} \frac{1}{K} (d_{ik} - \hat{d}_{ik})^2,
\end{equation}
where $K=128$ is the dimension of one output descriptor, $d_{ik}$ is the $k$-th dimension feature at point $i$ outputted by $\mathcal{N}_{\text{D}}$, $\hat{d}_{ik}$ is the counterpart outputted by HardNet.

\subsection{Multi-task Learning}

As our model complete two tasks simultaneously, it is a typical multi-task learning problem to train the model. Under deep learning framework, it is easy to perform this process. We can simply combine the two loss functions as
\begin{equation}
\label{eq:loss:final}
\mathcal{L} = \lambda_{\text{M}} \mathcal{L}_{\text{M}} + \lambda_{\text{D}} \mathcal{L}_{\text{D}},
\end{equation}
where $\lambda_{\text{M}}$ and $\lambda_{\text{D}}$ are the parameters to adjust the weight of each loss. In our experiment, we set $\lambda_{\text{M}} = 1, \lambda_{\text{D}} = 1$ empirically. With the multi-task learning strategy, we can simultaneously optimize the backbone network and two task-specific branches. One problem for multi-task learning is that the tasks may conflict with each other, which lead to an unsatisfied performance on some tasks. In this paper, this phenomenon has not appeared, which indicates that it is reasonable to integrate these two tasks into one model to save computational cost.

The complete training process can be summarized as follows. As demonstrated in Fig. \ref{fig:pipeline}, we first generate a mini-batch training sample from the training set and feed them into our model. Then our model outputs the motion attribute probabilities and descriptors. For the motion attribute, the ground truth is generated from semantic segmentation annotations. For descriptor, we take the output of HardNet as the supervision signal. The model can be optimized via minimizing the loss function \eqref{eq:loss:final} using Adam optimization algorithm. We set the batch size as $16$, set the initial learning rate as $l_{0} = 0.01$ and gradually reduce it after the $e$-th epoch as 
\begin{equation}
\label{eq:lr:update}
l_{e} = l_{0} \times b^{\frac{e}{E}},
\end{equation}
where $E = 100$ is the total number of epochs, $b = 0.01$ is the factor to control the rate of decay. To avoid over-fitting, we also set the weight decay as $1e-6$ in all experiments.

\begin{figure}
\center
  \includegraphics[width=7cm]{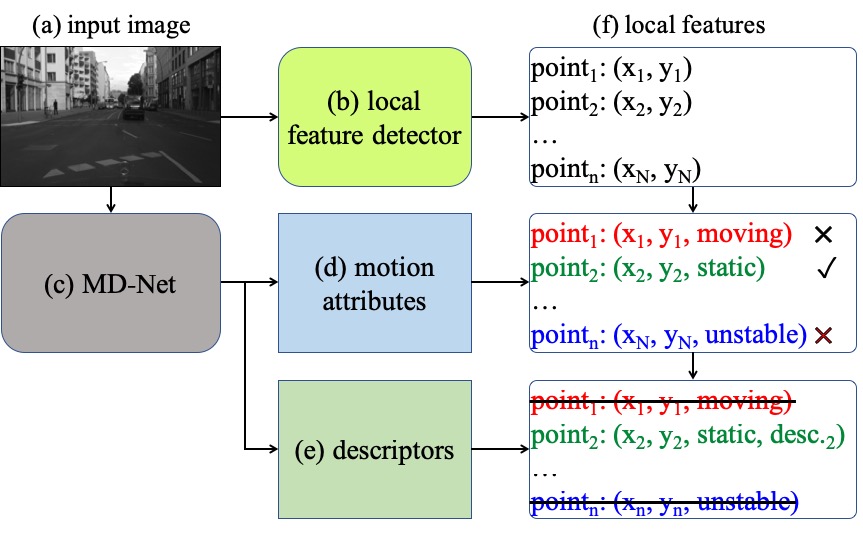}
  \caption{The process to integrate our model with a local feature detector.}
  \label{fig:slam:pipeline}
\end{figure}

\section{Association with localization System}
\label{sec:loc}
We integrate our model with a localization system to verify its effectiveness. As illustrated in Fig. \ref{fig:slam:pipeline}, we combine our model with a local feature detector. The detected feature points are filtered according to its motion attribute. Only static points are reserved for the following steps and the other points are discarded. Subsequently, the descriptor of each reserved point can be obtained from the outputs of the branch $\mathcal{N}_{\text{D}}$. Note that, as our model down-samples the resolution, we up-sample the results using bilinear interpolation. 

The localization algorithm used to test the proposed local feature pipeline is a sliding-window based visual-inertial SLAM method. Visual-inertial SLAM is widely used recently by utilizing the complementary properties of both cameras and inertial measurement units (IMU) to greatly enhance the localization performance~\cite{li2013high,li2013optimization,qin2017vins,zhang2019large,leutenegger2015keyframe}. The exact implementation follows that of~\cite{zhang2019large}, with the following changes. 
Firstly and most importantly, the proposed local feature pipeline is used instead of FREAK in~\cite{zhang2019large}.
Secondly, we do not use odometry measurements during our tests. As a result, the cost function does not contain the odometry terms. Due to the same reason, the poses are expressed with respect to IMU frame, and IMU to odometry extrinsic parameters are ignored. Lastly, the pose integration and keyframe selection policy are purely based on IMU integration.

\begin{figure*}
\center
  \includegraphics[width=15cm]{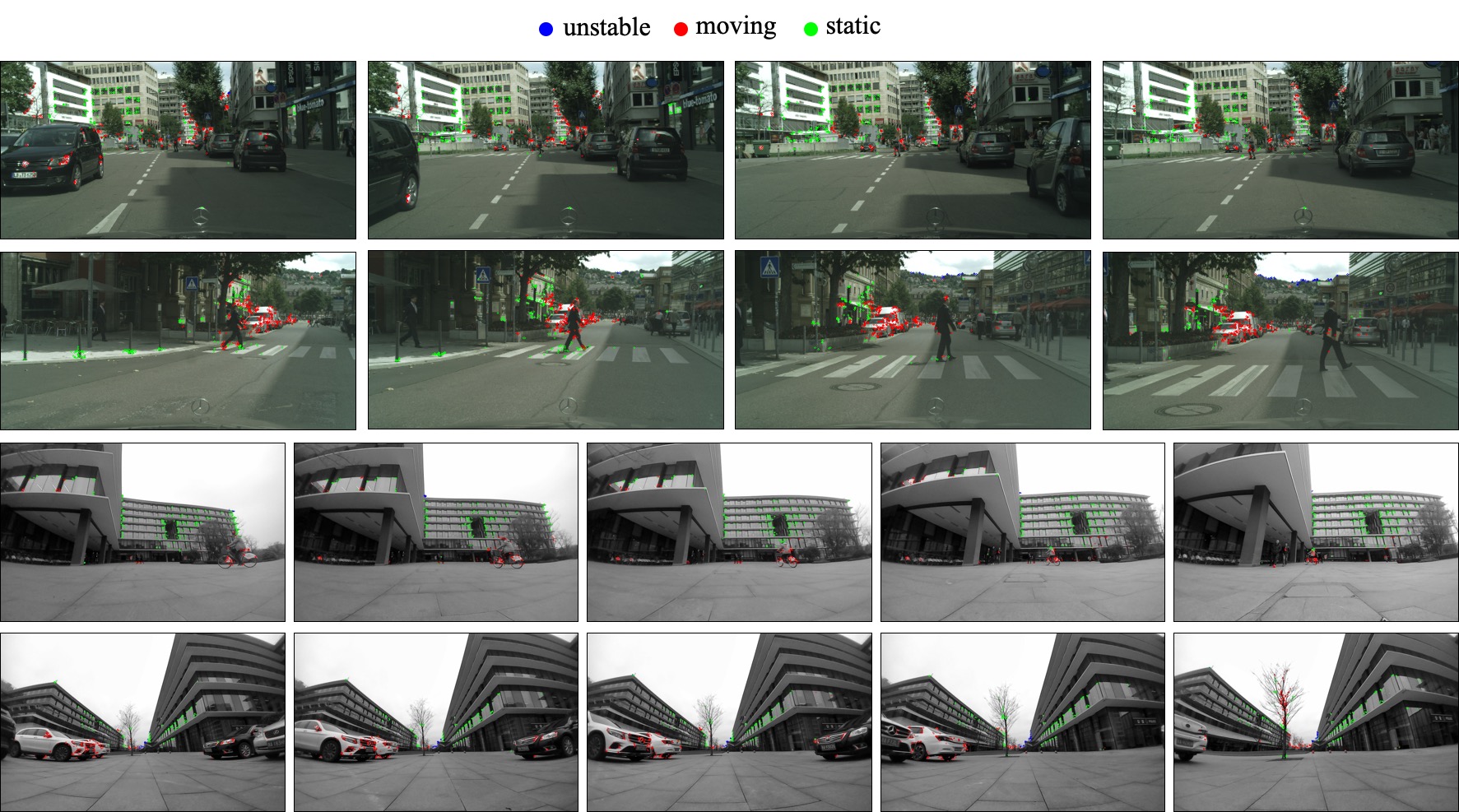}
  \caption{Motion attribute estimation results. Top two rows are results on Cityscapes dataset. Bottom two rows are results on the data collected in Alibaba campus, which are not in the training set.}
  \label{fig:motion:attribute:result}
\end{figure*}

\section{Experiments and Comparisons}

In this section, we first introduce experimental details to train our MD-Net and evaluate our motion attribute estimation results, our descriptors, and localization with our model in turn.

\begin{table}
\caption{The correspondence between motion attribute and semantic class in Cityscapes dataset.}
\label{tab:classes:motion}
\begin{center}
\begin{tabular}{ @{\hspace{1mm}} l @{\hspace{3mm}} l @{\hspace{1mm}} }
\hline \hline
Motion attribute  &  Semantic class \\
\hline
unstable  & sky, vegetation, terrain \\
\hline
moving   & human, vehicle, static, dynamic, traffic light \\
\hline
\multirow{2}{*}{static}  & ground, flat, building, wall, fence, guard rail, bridge,  \\
 &  tunnel, pole, pole group, traffic sign \\
\hline \hline
\end{tabular}
\end{center}
\end{table}

\subsection{Motion Attribute Estimation Results}

To train our MD-Net, we use the Cityscapes dataset \cite{Cordts2016Cityscapes}, which is with finely annotated semantic segmentation ground truth. This dataset is collected from various urban street scenes and consists of a training set with $2993$ images and a validation set with $503$ images. Before using the annotations, we transform the initial semantic classes into $3$ motion attribute as in Tab. \ref{tab:classes:motion}. The final model is obtained after trained on the training set. We implement our training process using PyTorch \cite{PyTorchNIPS2017} on a PC with an NVIDIA 1080 Ti GPU.

As shown in the third row of Tab. \ref{tab:iou:motion}, we can see that our model can accurately predict the motion attribute with a mean IoU as $76.2$. To verify the effectiveness of the class-reweighting strategy, we perform an ablation experiment by removing this strategy whose results are shown in the second row. The proportion of each class is also presented in the fourth row. We can see that this strategy can markedly improve the performance of the class with a small size, \ie~$5.4\%$ on unstable class, and increases the mean IoU about $2\%$. As shown in the top two rows of Fig. \ref{fig:motion:attribute:result}, our model also achieves good visual results. Moreover, to verify the generalization of our model, we also applied our model on some data from our work park. The results are shown in the bottom 2 rows of Fig. \ref{fig:motion:attribute:result}, which show that our model can well generalize to other data.

\begin{table}
\caption{IoU of Motion attribute on the Cityscapes validation dataset.}
\label{tab:iou:motion}
\small
\begin{center}
\begin{tabular}{ @{\hspace{2mm}} l @{\hspace{3mm}} c @{\hspace{3mm}} c @{\hspace{3mm}} c @{\hspace{3mm}} c @{\hspace{2mm}} }
\hline \hline
 & unstable & moving & static & mean \\
\hline
Our w/o re-weighting  &  62.9           & \textbf{74.6}        & 85.3        & 74.3 \\
Our                   &  \textbf{68.3}  & 74.5             & \textbf{85.7} & \textbf{76.2} \\
\hline
Proportion & 2.7\% & 35.8\% & 61.5\% & - \\
\hline \hline
\end{tabular}
\end{center}
\end{table}

\begin{figure*}
\center
  \includegraphics[width=15cm]{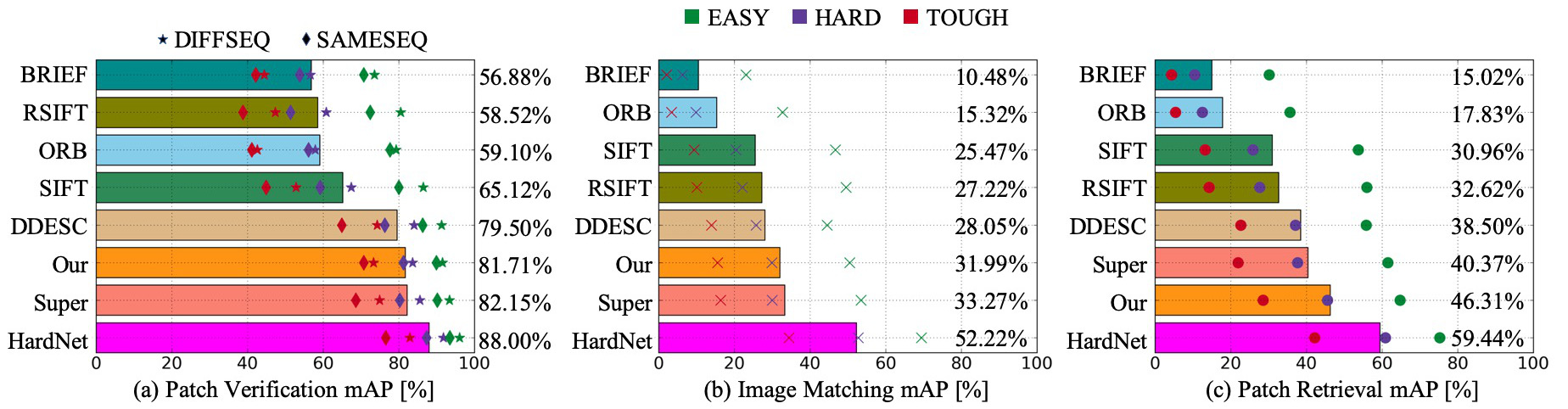}
  \caption{The performance of 8 different descriptors on HPatches.}
  \label{fig:descriptor:performance}
\end{figure*}

\subsection{Performance of the Descriptor}

To evaluate our descriptors, we use the HPatches to perform patch verification, matching and retrieval tasks on the FULL split. The details about the tasks and evaluation protocols can be found in \cite{HPatches_2017_cvpr}. The quantitative results are demonstrated in Fig. \ref{fig:descriptor:performance}. For a fair comparison, we direct use the pre-computed results of SIFT \cite{SIFT_2004_ijcv}, RootSIFT, BRIEF \cite{brief_2010_eccv}, ORB \cite{ORB_2011_iccv},  and DDESC \cite{ddesc_2015_iccv} released by HPatches. The results of SuperPoint \cite{superpoint_2018_cvpr} and HardNet \cite{Mishchuk_2017_nips} are calculated using the released models by the authors. We can see that our model can successfully mimic HardNet and achieve comparable results on verification and matching tasks compared with previous FCN based method SuperPoint, and outperforms it on the retrieval task with a large margin. The results also indicate that HardNet, a patch-based method, can obtain better performance than FCN based methods, which would be a future work of FCN based methods.

\subsection{Performance in Vision based Localization}

In this section, we present the results of vision-based localization using the proposed features from MD-Net.
In our experiments, the sensor suite consists of a MYNT camera and a BOSCH BMI160 IMU. The camera captured images at 10Hz, and the IMU measurements were provided at 200Hz.
The temporal, intrinsic, and extrinsic parameters of those sensors were calibrated offline using the method in \cite{li2014high}.
During the data collection process, the sensor suite was mounted on ground vehicles (robots or cars).

The first experiment is to evaluate the inlier feature ratio via two-view based RANSAC algorithm at three different testing environments, \ie, normal scene, with pedestrians, and with a couple of slowly moving vehicles. Tab. \ref{tab:inlier:ratio} shows the RANSAC inlier ratio for different local feature algorithms, which is computed by averaging all image pairs used in localization. For the compared local feature algorithms, the {\em same} detector was used for focusing on the descriptor comparison. Results in Tab. \ref{tab:inlier:ratio} demonstrate that the proposed algorithm is able to reach the best RANSAC inlier ratio, meaning that features corresponding to moving and unstable objects can be pre-removed. The proposed algorithm also outperforms SuperPoint in all cases.

\begin{table}[t]
\caption{Inlier ratio on the data from 3 different scenes.}
\small
\label{tab:inlier:ratio}
\begin{center}
\begin{tabular}{ @{\hspace{2mm}} l @{\hspace{4mm}} l @{\hspace{4mm}} c @{\hspace{4mm}} c @{\hspace{4mm}} c @{\hspace{4mm}} c @{\hspace{2mm}}}
\hline \hline
Detector  &  Descriptor & Scene1 & Scene2 & Scene3 & Mean \\
\hline
FAST & FREAK & 87.6 & 87.1 & 84.1 &  86.2 \\ 
FAST & SuperPoint & 88.8 & 88.4 & 90.2 & 89.1 \\
FAST & Our & \textbf{91.2} & \textbf{93.3} & \textbf{92.2} &  \textbf{92.2} \\  
\hline \hline
\end{tabular}
\end{center}
\end{table}

\begin{figure}[t]
\center
  \includegraphics[width=\columnwidth]{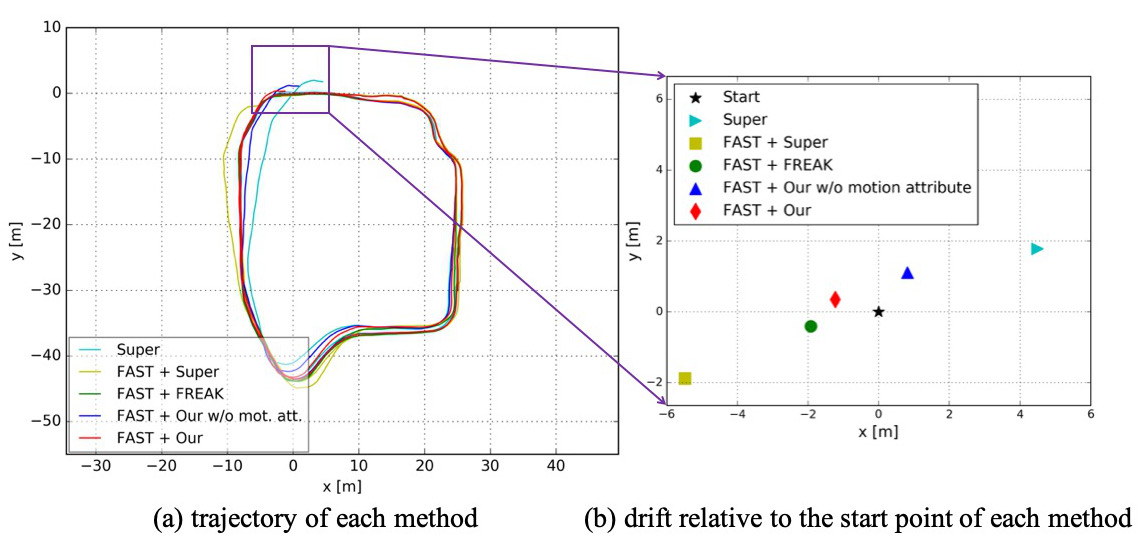}
  \caption{Localization trajectory and drift. }
  \label{fig:slam:performance}
\end{figure}

\begin{table}[t]
\caption{Localization drift.}
\small
\label{tab:mapping:drift}
\begin{center}
\begin{tabular}{ @{\hspace{2mm}} l @{\hspace{4mm}} l @{\hspace{4mm}} c @{\hspace{4mm}} c @{\hspace{2mm}}}
\hline \hline
Detector  &  Descriptor & Drift (x, y) [m] & Drift [m] \\
\hline
FAST  & FREAK & (-1.9, -0.4) & 1.94 \\
FAST  & SuperPoint & (-5.4, -1.8) & 5.69 \\
SuperPoint & SuperPoint & (4.4, 1.8) & 4.75 \\
\hline
FAST  & Our w/o mot. att. & (0.8, 1.1) & 1.36 \\
FAST  & Our & (-1.2, 0.3) & \textbf{1.23} \\
\hline \hline
\end{tabular}
\end{center}
\end{table}

The second and the third experiments are to show the performance of the proposed visual localization algorithm using our local feature method, in two different environments. In the second test, we collected a dataset around a local restaurant by a ground robot, whose trajectory started and stopped at the {\em exactly} the same location. This allows us to compute the `final drift' as an error metric.
Tab. \ref{tab:mapping:drift} and Fig. \ref{fig:slam:performance} show the final drifts by using different feature detectors and descriptors but under the same localization method. A couple of observations can be made from the result. Firstly, by explicitly modeling motion attribute, the localization error is reduced compared to the alternative method without the attribute. This is consistent with our claim that motion attribute is able to filter out more `bad' features. Additionally, we note that our proposed method achieves the best precision, compared to all other methods, including SuperPoint families and classic FAST/FREAK combination.
These results show that the proposed method is the best one for performing vision-based localization, at least among the experiments we conducted.

\begin{table}[t]
\caption{The root-mean-squared-error[m] between the result and the ground-truth trajectories.}
\small
\label{tab:mapping:error}
\begin{center}
\begin{tabular}{ @{\hspace{2mm}} l @{\hspace{4mm}} l @{\hspace{4mm}} c @{\hspace{4mm}} c @{\hspace{2mm}} }
\hline \hline
Detector  &  Descriptor & On series1 & On series2 \\
\hline
FAST  & FREAK & 2.87 & 27.06 \\
FAST  & SuperPoint & 10.02 & Failed \\
SuperPoint & SuperPoint & 7.96 & Failed \\
\hline
FAST  & Our w/o mot. att. & 7.82 & 20.86 \\
FAST  & Our & \textbf{1.10} & \textbf{7.25} \\
\hline \hline
\multicolumn{2}{c}{Series length} & 279.1 & 880.8 \\
\hline \hline
\end{tabular}
\end{center}
\end{table}

\begin{figure}[t]
\centering	
  \includegraphics[width=0.9\columnwidth]{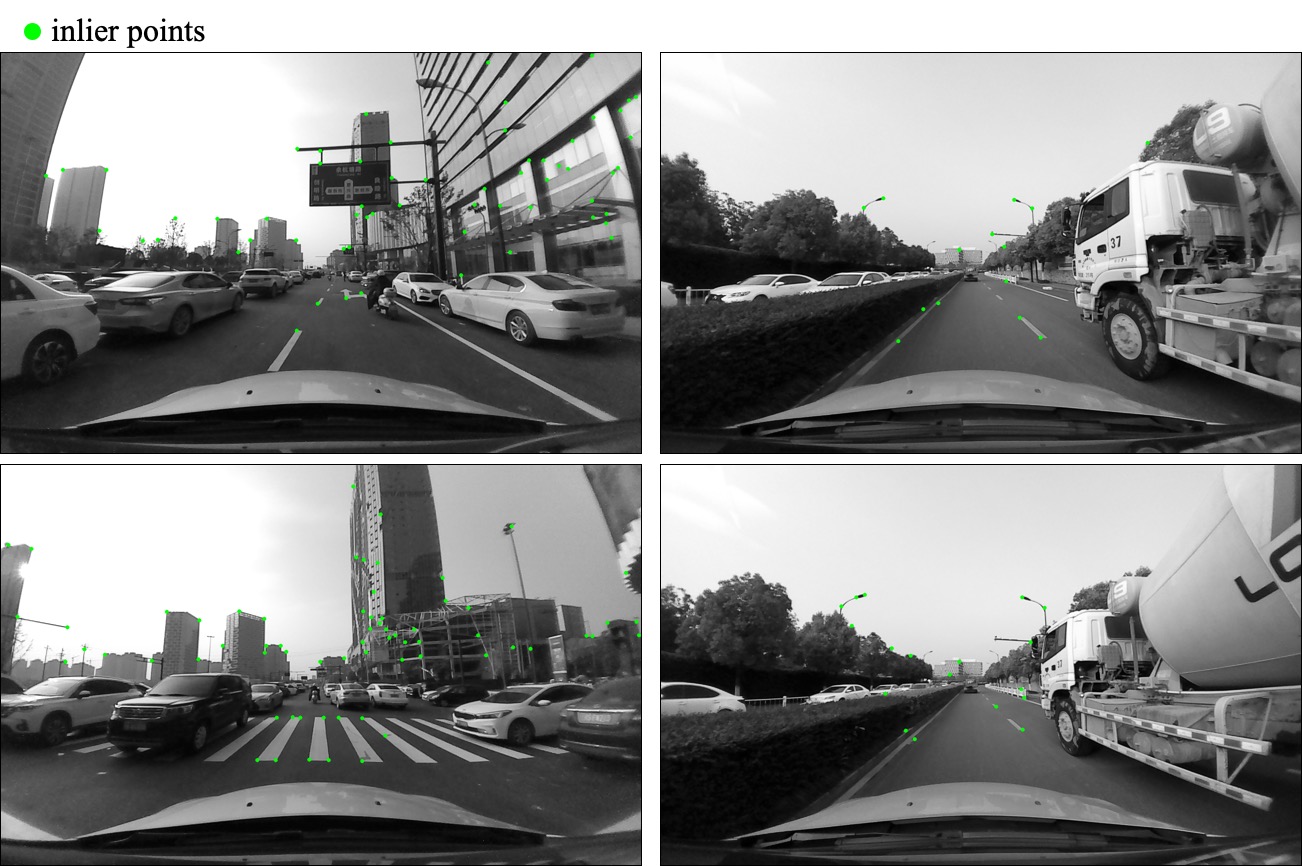}
  \caption{Representative images captured during the data collection of the urban street view datasets.}
  \label{fig:image2}
\end{figure}

In the last experiment, we collected two series of urban street view datasets by mounting our sensors along with RTK-GPS on top of a car. The representative images captured during the data collection process are shown in Fig.~\ref{fig:image2}. The readings from RTK-GPS are taken as the ground-truth pose. For quantitative evaluation, the computed trajectory is aligned to the ground-truth using Umeyama's method \cite{umeyama1991least}. The root-mean-square errors of each method can be found in Tab.~\ref{tab:mapping:error}. We can see that our method obtains the lowest error among several comparison methods. We also note that the dataset is extremely challenging, where some images are full of or dominated by moving vehicles (see Fig.~\ref{fig:image2}). However, by filtering out improper features by motion attribute, the proposed method can still achieve high-precision localization.

%


\section{Discussion and Conclusion}

In this paper, we design a fully convolutional network MD-Net to perform motion attribute estimation and feature description simultaneously. MD-Net has three modules: the backbone network $\mathcal{N}_{\text{B}}$ to extract shared features from the input image, the motion attribute branch $\mathcal{N}_{\text{M}}$ to estimate motion attribute, the description branch $\mathcal{N}_{\text{D}}$ to extract distinctive descriptor. We further integrate the proposed method into a visual-inertial localization system to perform high-precision pose estimation. Experimental results demonstrate that the proposed method can improve the performance and outperforms previous similar algorithms, especially in complicated dynamic environments when multiple moving objects exist. 
The limitation of this paper is that the proposed method still relies on a third-party feature detector, which will be integrated into our FCN model in the future to enhance the performance further. 

%

\section*{ACKNOWLEDGMENT}

We would like to thank Dongsheng Hong for his invaluable help. This work was supported by grants from the National Key R\&D Program of China (2017YFB1002400), the National Natural Science Foundation of China (61672072, U1611461 and 61825101), Beijing Nova Program (Z181100006218063), and China Postdoctoral Science Foundation (2018M641110).



\bibliographystyle{IEEEtran}
\bibliography{ref.bib}

\begin{thebibliography}{10}
\providecommand{\url}[1]{#1}
\csname url@rmstyle\endcsname
\providecommand{\newblock}{\relax}
\providecommand{\bibinfo}[2]{#2}
\providecommand\BIBentrySTDinterwordspacing{\spaceskip=0pt\relax}
\providecommand\BIBentryALTinterwordstretchfactor{4}
\providecommand\BIBentryALTinterwordspacing{\spaceskip=\fontdimen2\font plus
\BIBentryALTinterwordstretchfactor\fontdimen3\font minus
  \fontdimen4\font\relax}
\providecommand\BIBforeignlanguage[2]{{%
\expandafter\ifx\csname l@#1\endcsname\relax
\typeout{** WARNING: IEEEtran.bst: No hyphenation pattern has been}%
\typeout{** loaded for the language `#1'. Using the pattern for}%
\typeout{** the default language instead.}%
\else
\language=\csname l@#1\endcsname
\fi
#2}}

\bibitem{Song2016TMM}
Y.~Song, X.~Chen, X.~Wang, Y.~Zhang, and J.~Li, ``{6-DOF Image Localization
  from Massive Geo-tagged Reference Images},'' \emph{IEEE Transactions on
  Multimedia}, vol.~18, no.~8, pp. 1542--1554, 2016.

\bibitem{li2014high}
M.~Li, H.~Yu, X.~Zheng, and A.~I. Mourikis, ``High-fidelity sensor modeling and
  self-calibration in vision-aided inertial navigation,'' in \emph{ICRA}, May
  2014, pp. 409--416.

\bibitem{li2013high}
M.~Li and A.~I. Mourikis, ``High-precision, consistent ekf-based
  visual-inertial odometry,'' \emph{IJRR}, vol.~32, no.~6, pp. 690--711, 2013.

\bibitem{concha2015dpptam}
A.~Concha and J.~Civera, ``Dpptam: Dense piecewise planar tracking and mapping
  from a monocular sequence,'' in \emph{IEEE/RSJ International Conference on
  Intelligent Robots and Systems}, 2015, pp. 5686--5693.

\bibitem{Huang_2018_ICRA}
R.~{Huang}, D.~{Zou}, R.~{Vaughan}, and P.~{Tan}, ``{Active Image-Based
  Modeling with a Toy Drone},'' in \emph{ICRA}, 2018.

\bibitem{Zhu_2018_CVPR}
S.~Zhu, R.~Zhang, L.~Zhou, T.~Shen, T.~Fang, P.~Tan, and L.~Quan, ``{Very
  Large-Scale Global SfM by Distributed Motion Averaging},'' in \emph{CVPR},
  2018.

\bibitem{Mur_2017_orb_slam2}
R.~Mur{-}Artal and J.~D. Tard{\'{o}}s, ``{{ORB-SLAM2:} An Open-Source {SLAM}
  System for Monocular, Stereo, and {RGB-D} Cameras},'' \emph{{IEEE}
  Transactions on Robotics}, vol.~33, no.~5, pp. 1255--1262, 2017.

\bibitem{schneider2018maplab}
T.~Schneider, M.~Dymczyk, M.~Fehr, K.~Egger, S.~Lynen, I.~Gilitschenski, and
  R.~Siegwart, ``maplab: An open framework for research in visual-inertial
  mapping and localization,'' \emph{IEEE Robotics and Automation Letters},
  2018.

\bibitem{li2013optimization}
M.~Li and A.~I. Mourikis, ``Optimization-based estimator design for
  vision-aided inertial navigation,'' in \emph{Robotics: Science and Systems},
  2013, pp. 241--248.

\bibitem{SIFT_2004_ijcv}
D.~G. Lowe, ``{Distinctive Image Features from Scale-Invariant Keypoints},''
  \emph{IJCV}, vol.~60, no.~2, pp. 91--110, 2004.

\bibitem{ORB_2011_iccv}
E.~Rublee, V.~Rabaud, K.~Konolige, and G.~R. Bradski, ``{{ORB:} An efficient
  alternative to {SIFT} or {SURF}},'' in \emph{ICCV}, 2011.

\bibitem{FREAK_2012_cvpr}
A.~Alahi, R.~Ortiz, and P.~Vandergheynst, ``{{FREAK:} Fast Retina Keypoint},''
  in \emph{CVPR}, 2012.

\bibitem{Ren_2017_faster_rcnn}
S.~Ren, K.~He, R.~B. Girshick, and J.~Sun, ``{Faster {R-CNN:} Towards Real-Time
  Object Detection with Region Proposal Networks},'' \emph{{IEEE} Trans.
  Pattern Anal. Mach. Intell.}, vol.~39, no.~6, pp. 1137--1149, 2017.

\bibitem{Shelhamer_2017_FCN}
E.~Shelhamer, J.~Long, and T.~Darrell, ``{Fully Convolutional Networks for
  Semantic Segmentation},'' \emph{{IEEE} Trans. Pattern Anal. Mach. Intell.},
  vol.~39, no.~4, pp. 640--651, 2017.

\bibitem{TILDE_2015_cvpr}
Y.~Verdie, K.~M. Yi, P.~Fua, and V.~Lepetit, ``{{TILDE:} {A} Temporally
  Invariant Learned DEtector},'' in \emph{CVPR}, 2015.

\bibitem{LIFT_2016_eccv}
K.~M. Yi, E.~Trulls, V.~Lepetit, and P.~Fua, ``{{LIFT:} Learned Invariant
  Feature Transform},'' in \emph{ECCV}, 2016.

\bibitem{Tian_2017_cvpr}
Y.~Tian, B.~Fan, and F.~Wu, ``{L2-Net: Deep Learning of Discriminative Patch
  Descriptor in Euclidean Space},'' in \emph{CVPR}, 2017.

\bibitem{Mishchuk_2017_nips}
A.~Mishchuk, D.~Mishkin, F.~Radenovic, and J.~Matas, ``{Working Hard to Know
  Your Neighbor's Margins: Local Descriptor Learning Loss},'' in \emph{Advances
  in Neural Information Processing Systems (NeurIPS)}, 2017.

\bibitem{superpoint_2018_cvpr}
D.~DeTone, T.~Malisiewicz, and A.~Rabinovich, ``{SuperPoint: Self-Supervised
  Interest Point Detection and Description},'' in \emph{CVPR - Workshops},
  2018.

\bibitem{MVG_2004_book}
R.~I. Hartley and A.~Zisserman, \emph{Multiple View Geometry in Computer
  Vision}, 2nd~ed.\hskip 1em plus 0.5em minus 0.4em\relax Cambridge University
  Press, ISBN: 0521540518, 2004.

\bibitem{lynen2015get}
S.~Lynen, T.~Sattler, M.~Bosse, J.~A. Hesch, M.~Pollefeys, and R.~Siegwart,
  ``Get out of my lab: Large-scale, real-time visual-inertial localization.''
  in \emph{Robotics: Science and Systems}, 2015.

\bibitem{zhang2019large}
M.~Zhang, Y.~Chen, and M.~Li, ``Vision-aided localization for ground robots,''
  in \emph{IEEE/RSJ International Conference on Intelligent Robots and Systems
  (IROS)}, 2019.

\bibitem{mask_slam_2018_cvpr}
M.~Kaneko, K.~Iwami, T.~Ogawa, T.~Yamasaki, and K.~Aizawa, ``{Mask-SLAM: Robust
  Feature-Based Monocular {SLAM} by Masking Using Semantic Segmentation},'' in
  \emph{CVPR - Workshops}, 2018.

\bibitem{Naseer_2017_icra}
T.~Naseer, G.~L. Oliveira, T.~Brox, and W.~Burgard, ``{Semantics-aware Visual
  Localization Under Challenging Perceptual Conditions},'' in \emph{ICRA},
  2017.

\bibitem{rcnn_2014_cvpr}
R.~B. Girshick, J.~Donahue, T.~Darrell, and J.~Malik, ``{Rich Feature
  Hierarchies for Accurate Object Detection and Semantic Segmentation},'' in
  \emph{CVPR}, 2014.

\bibitem{superpoint_v2_2018_arxiv}
D.~DeTone, T.~Malisiewicz, and A.~Rabinovich, ``{Self-Improving Visual
  Odometry},'' \emph{CoRR}, vol. abs/1812.03245, 2018.

\bibitem{Szegedy_2013_nips}
C.~Szegedy, A.~Toshev, and D.~Erhan, ``{Deep Neural Networks for Object
  Detection},'' in \emph{Advances in Neural Information Processing Systems
  (NeurIPS)}, 2013.

\bibitem{Harris_1988_AVC}
C.~G. Harris and M.~Stephens, ``{A Combined Corner and Edge Detector},'' in
  \emph{Alvey Vision Conference {(AVC)}}, 1988.

\bibitem{FAST_2006_eccv}
E.~Rosten and T.~Drummond, ``{Machine Learning for High-Speed Corner
  Detection},'' in \emph{{ECCV}}, 2006.

\bibitem{Lindeberg_1998_ijcv}
T.~Lindeberg, ``{Edge Detection and Ridge Detection with Automatic Scale
  Selection},'' \emph{IJCV}, vol.~30, no.~2, pp. 117--156, 1998.

\bibitem{MSER_2002_bmvc}
J.~Matas, O.~Chum, M.~Urban, and T.~Pajdla, ``{Robust Wide Baseline Stereo from
  Maximally Stable Extremal Regions},'' in \emph{{BMVC}}, 2002.

\bibitem{SURF_2006_eccv}
H.~Bay, T.~Tuytelaars, and L.~J.~V. Gool, ``{{SURF:} Speeded Up Robust
  Features},'' in \emph{ECCV}, 2006.

\bibitem{Zhu_2006_cvpr}
Q.~Zhu, M.~Yeh, K.~Cheng, and S.~Avidan, ``{Fast Human Detection Using a
  Cascade of Histograms of Oriented Gradients},'' in \emph{CVPR}, 2006.

\bibitem{brief_2010_eccv}
M.~Calonder, V.~Lepetit, C.~Strecha, and P.~Fua, ``{{BRIEF:} Binary Robust
  Independent Elementary Features},'' in \emph{ECCV}, 2010.

\bibitem{Savinov2017}
N.~Savinov, A.~Seki, L.~Ladicky, T.~Sattler, and M.~Pollefeys,
  ``{Quad-networks: unsupervised learning to rank for interest point
  detection},'' in \emph{CVPR}, 2017.

\bibitem{Zhang2017}
X.~Zhang, F.~X. Yu, S.~Karaman, and S.-F. Chang, ``{Learning Discriminative and
  Transformation Covariant Local Feature Detectors},'' in \emph{CVPR}, 2017.

\bibitem{Brown_2007_ijcv}
M.~Brown and D.~G. Lowe, ``{Automatic Panoramic Image Stitching using Invariant
  Features},'' \emph{IJCV}, vol.~74, no.~1, pp. 59--73, 2007.

\bibitem{HPatches_2017_cvpr}
V.~Balntas, K.~Lenc, A.~Vedaldi, and K.~Mikolajczyk, ``{HPatches: {A} Benchmark
  and Evaluation of Handcrafted and Learned Local Descriptors},'' in
  \emph{CVPR}, 2017.

\bibitem{PS_Dataset_2018_arxiv}
R.~Mitra, N.~Doiphode, U.~Gautam, S.~Narayan, S.~Ahmed, S.~Chandran, and
  A.~Jain, ``{A Large Dataset for Improving Patch Matching},'' \emph{CoRR},
  vol. abs/1801.01466, 2018.

\bibitem{Luo2018}
Z.~Luo, T.~Shen, L.~Zhou, S.~Zhu, R.~Zhang, Y.~Yao, T.~Fang, and L.~Quan,
  ``{GeoDesc: Learning Local Descriptors by Integrating Geometry
  Constraints},'' in \emph{ECCV}, 2018.

\bibitem{Ono2018}
Y.~Ono, E.~Trulls, P.~Fua, and K.~M. Yi, ``{LF-Net: Learning Local Features
  from Images},'' in \emph{Advances in Neural Information Processing Systems
  (NeurIPS)}, 2018.

\bibitem{Ioffe_Szegedy_2015_batchnorm}
S.~Ioffe and C.~Szegedy, ``{Batch Normalization: Accelerating Deep Network
  Training by Reducing Internal Covariate Shift},'' in \emph{ICML}, 2015.

\bibitem{Nair_Hinton_2010_relu}
V.~Nair and G.~E. Hinton, ``{Rectified Linear Units Improve Restricted
  Boltzmann Machines},'' in \emph{ICML}, 2010.

\bibitem{Cordts2016Cityscapes}
M.~Cordts, M.~Omran, S.~Ramos, T.~Rehfeld, M.~Enzweiler, R.~Benenson,
  U.~Franke, S.~Roth, and B.~Schiele, ``{The Cityscapes Dataset for Semantic
  Urban Scene Understanding},'' in \emph{CVPR}, 2016.

\bibitem{Li_2018_arxiv}
J.~Li, Y.~Song, J.~Zhu, L.~Cheng, Y.~Su, L.~Ye, P.~Yuan, and S.~Han,
  ``{Learning from Large-scale Noisy Web Data with Ubiquitous Reweighting for
  Image Classification},'' \emph{CoRR}, vol. abs/1811.00700, 2018.

\bibitem{Song_2017_iccv}
Y.~Song, X.~Chen, J.~Li, and Q.~Zhao, ``{Embedding 3D Geometric Features for
  Rigid Object Part Segmentation},'' in \emph{ICCV}, 2017.

\bibitem{qin2017vins}
T.~Qin, P.~Li, and S.~Shen, ``Vins-mono: A robust and versatile monocular
  visual-inertial state estimator,'' \emph{IEEE Transactions on Robotics},
  vol.~34, no.~4, pp. 1004--1020, 2018.

\bibitem{leutenegger2015keyframe}
S.~Leutenegger, S.~Lynen, M.~Bosse, R.~Siegwart, and P.~Furgale,
  ``Keyframe-based visual--inertial odometry using nonlinear optimization,''
  \emph{IJRR}, vol.~34, no.~3, pp. 314--334, 2015.

\bibitem{PyTorchNIPS2017}
A.~Paszke, S.~Gross, S.~Chintala, G.~Chanan, E.~Yang, Z.~DeVito, Z.~Lin,
  A.~Desmaison, L.~Antiga, and A.~Lerer, ``Automatic differentiation in
  pytorch,'' in \emph{Advances in Neural Information Processing Systems
  (NeurIPS) - Workshop}, 2017.

\bibitem{ddesc_2015_iccv}
E.~Simo{-}Serra, E.~Trulls, L.~Ferraz, I.~Kokkinos, P.~Fua, and
  F.~Moreno{-}Noguer, ``{Discriminative Learning of Deep Convolutional Feature
  Point Descriptors},'' in \emph{ICCV}, 2015.

\bibitem{umeyama1991least}
S.~Umeyama, ``Least-squares estimation of transformation parameters between two
  point patterns,'' \emph{IEEE Trans. Pattern Anal. Mach. Intell.}, no.~4, pp.
  376--380, 1991.

\end{thebibliography}

\end{document}